\PassOptionsToPackage{table}{xcolor}
\documentclass[10pt,twocolumn,letterpaper]{article}

\usepackage[pagenumbers]{wacv}%

\usepackage[T1]{fontenc}
\usepackage[utf8]{inputenc}

\definecolor{wacvblue}{rgb}{0.21,0.49,0.74}
\usepackage[pagebackref,breaklinks,colorlinks,allcolors=wacvblue]{hyperref}
\usepackage[acronym,nomain,nonumberlist]{glossaries}
\usepackage{multirow}
\usepackage{tabularx}
\usepackage{array}
\usepackage{setspace}
\usepackage{iftex}
\ifPDFTeX\usepackage[accsupp]{axessibility}\fi

\newacronym{od}{OD}{Object Detection}
\newacronym{dl}{DL}{Deep Learning}
\newacronym{fsl}{FSL}{Few-shot Learning}
\newacronym{vfm}{VFM}{Visual Foundation Model}
\newacronym{sam}{SAM}{Segment Anything Model}
\newacronym{rgb}{RGB}{red--green--blue}
\newacronym{uav}{UAV}{Unmanned Aerial Vehicle}
\newacronym{dino}{DINOv2}{self-DIstillation with NO labels}
\newacronym{mae}{MAE}{Mean Absolute Error}
\newacronym{iou}{IoU}{Intersection over Union}
\newacronym{radio}{AM-RADIO}{Agglomerative Model--Reduce All Domains Into One}
\newacronym{geco2}{GeCo2}{Generalized-Scale Object Counting}
\newacronym{petdino}{PET-DINO}{Prompt-Enriched Grounding DINO}
\def\wacvPaperID{19}%
\def\confName{WACV}
\def\confYear{2027}

\title{Decoding Frozen Vision Representations for Training-Free Few-Shot \\ Pineapple Detection}

\author{
Manfred Gonzalez-Hernandez$^{1, \dagger}$ \quad
Danny Xie-Li$^{2}$ \quad
Fabian Fallas-Moya$^{3}$ \quad
Camile Lendering$^{1}$ \\[0.5em]
Joaquín Figueira$^{1}$ \quad
Erkut Akdag$^{1}$ \quad
Giacomo D'Amicantonio$^{1}$ \quad
Egor Bondarev$^{1}$ \\[1em]
$^{1}$ Eindhoven University of Technology \quad
$^{2}$ University of Luxembourg \quad
$^{3}$ Universidad de Costa Rica \\[0.3em]
$^{\dagger}$ Corresponding author: {\tt\small  m.gonzalez.hernandez@tue.nl}
}

\usepackage{microtype}
\begin{document}
\maketitle
\begin{abstract}
Few-shot object detection in agricultural imagery is hampered by dense vegetation, repeated patterns, and scarce annotations. We present a training-free detector that turns dense features from frozen visual foundation models directly into detections, without fine-tuning, a learned head, or external proposals: a whitened foreground--background direction estimated from annotated support images scores every query patch, and the score field is decoded with an adaptive threshold and support-derived geometry. We also test whether predicted relative depth can be fed directly to RGB-pretrained frozen encoders. On aerial pineapple imagery, the detector outperforms PET-DINO and GeCo2 at every evaluated support size; at $K$=40, predicted depth alone reaches 92\% of the RGB performance with DINOv2, and RGB--depth fusion lowers counting error for both encoders. We release the Pineapple Plantation dataset (4,908 aerial frames, 1,995 annotated with 17,609 boxes, predicted depth, leakage-aware splits) and analyse transfer to two further agricultural datasets.\footnote{Code and benchmark are available at \url{https://github.com/imagine-laboratory/Dino-Radio}}
\end{abstract}

\section{Introduction}
\begin{figure}[t]
  \centering
  \includegraphics[width=0.97\linewidth]{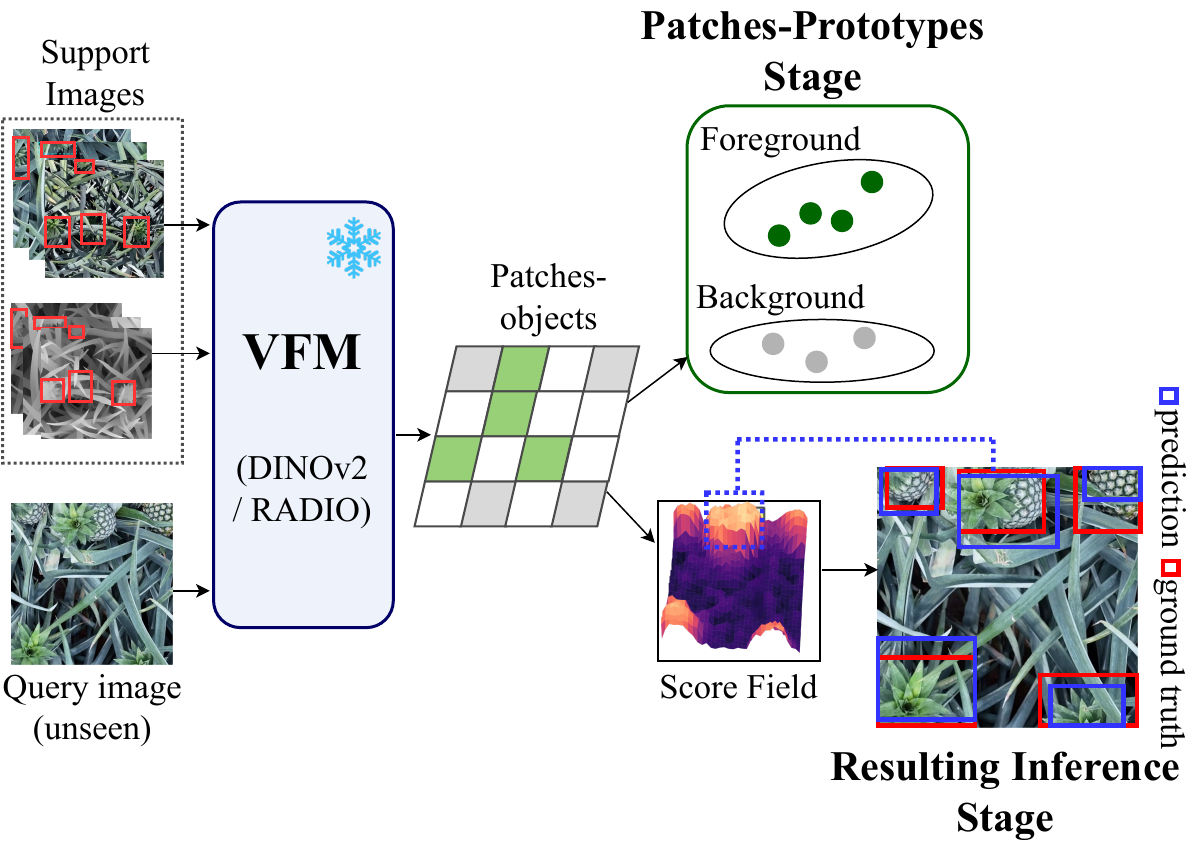}
    \caption{Training-free few-shot detection using frozen RGB+depth features. $K$ annotated support images define foreground/background patch prototypes whose similarity field localizes objects in unseen queries, without fine-tuning or learned detection heads. At $K=40$, our method achieves $1.5\times$ the \gls{iou} F1 of the strongest baseline, while RGB+depth fusion reduces count \gls{mae} for both encoders.}

    \label{fig:gaudeidiff_init}
\end{figure}

Accurate crop counts are important for production forecasting, loss estimation, and harvest planning in large-scale agriculture. Pineapple cultivation is a particularly relevant setting: large volumes of imagery can be collected through low-altitude \gls{uav} flights, but manually annotating every plant remains costly~\cite{kamilaris2018review,sa2016deepfruits}. Pineapple crowns are also difficult to localize because they are embedded in dense foliage, partially occluded, and visually similar to the surrounding leaves. At the studied commercial plantation (Upala Agricola, one of the world's largest exporters of fresh pineapple) in Costa Rica, counts are estimated by workers traversing selected field lines covering approximately 2--3\% of the plantation. Our \gls{uav} flights at approximately 2\,m were designed to emulate this sampling process while preserving the close-range view needed to distinguish individual crowns. Since reliable counting requires identifying individual plants rather than only estimating image-level density, we formulate the problem as few-shot \gls{od}: localize each pineapple crown from a small annotated support set and derive counts from the resulting detections.

Recent \glspl{vfm}, including \gls{dino}~\cite{oquab2023dinov2} and \gls{radio}~\cite{ranzinger2024amradio,heinrich2025radiov25}, provide transferable dense representations. Few-shot methods such as \gls{petdino}~\cite{Fu_2026_PETDINO} and \gls{geco2}~\cite{Pelhan_2026_GeCo2} exploit visual prompting for detection and counting, but rely primarily on \gls{rgb} appearance, which can be ambiguous in dense vegetation. Although frozen visual representations encode geometric information, whether predicted depth can improve detection when processed by \gls{rgb}-pretrained frozen encoders remains unclear.

We study whether frozen dense features can support few-shot pineapple detection and counting directly, while examining whether predicted depth provides complementary geometric information within the same frozen feature space. We evaluate \gls{dino}~\cite{oquab2023dinov2} and \gls{radio}~\cite{ranzinger2024amradio,heinrich2025radiov25}. From annotated support images, foreground and background patches are identified at multiple box-interior scales and summarized into a covariance-whitened prototype direction. Applying this direction to each query patch yields a spatial score field, which is separated with a support-selected two- or three-population Otsu criterion~\cite{otsu1979threshold} and converted into individual detections using support-derived geometry. No query annotations, gradient updates, learned detection head, or external proposal model are required.

Each aerial image may contain multiple crowns, so we define the few-shot setting in terms of \emph{support images} rather than individual object instances. Because \gls{rgb} appearance can be ambiguous under dense foliage, we evaluate \gls{rgb}, predicted relative depth, and \gls{rgb}+depth fusion to determine how much geometric information can be exploited by frozen \gls{rgb}-pretrained encoders for detection and counting. We test transfer to two additional agricultural datasets.

The main contributions are:

\begin{itemize}
    \item A training-free few-shot detector that converts frozen dense features directly into object detections and counts, without a learned head or external proposals.

    \item A study of predicted relative depth as input to \gls{rgb}-pretrained frozen encoders, including depth-only and \gls{rgb}--depth fusion without cross-modal adaptation.

    \item The Pineapple Plantation Dataset, comprising 4,908 aerial frames, including 1,995 annotated frames with 17,609 boxes, predicted depth, and leakage-aware splits.

    \item An evaluation showing improved pineapple counting and detection over \gls{petdino} and \gls{geco2}, scalability with additional support, domain-dependent transfer, and encoder-dependent benefits from \gls{rgb}--depth fusion.
\end{itemize}
\section{Related Work}

\textbf{Object Detection and Few-Shot Learning.}
Classical \gls{od} includes two-stage detectors such as Faster R-CNN~\cite{ren2015faster} and single-stage detectors such as YOLO~\cite{redmon2016yolo}, both typically requiring substantial annotation. Few-shot learning reduces this burden by adapting from few examples, as in feature reweighting~\cite{kang2019fewshot}, two-stage fine-tuning~\cite{wang2020frustratingly}, and prototype-based methods~\cite{snell2017prototypical}. For detection, however, class recognition must also be coupled with localization.

\textbf{Dense Representations and Foundation Models.}
Self-supervised transformers expose dense semantic and spatial structure~\cite{caron2021emerging}. LOST~\cite{simeoni2021lost} and TokenCut~\cite{wang2022tokencut} exploit this structure for unsupervised object discovery, while \gls{dino}~\cite{oquab2023dinov2}, \gls{radio}~\cite{ranzinger2024amradio}, and RADIOv2.5~\cite{heinrich2025radiov25} provide more general dense representations. These works motivate direct localization from frozen feature grids, but mainly address class-agnostic discovery rather than few-shot localization of a target class.

\textbf{Few-Shot and Training-Free Detection.}
FS-DETR~\cite{bulat2023fsdetr} and DE-ViT~\cite{zhang2025devit} use visual examples for few-shot detection without novel-class retraining. More recent approaches include \gls{petdino}~\cite{Fu_2026_PETDINO} and \gls{geco2}~\cite{Pelhan_2026_GeCo2}, while Matcher~\cite{liu2024matcher} and ProtoSAM~\cite{ayzenberg2024protosam} combine frozen features with segmentation mechanisms. Training-free detection has also been explored with external proposal models such as \gls{sam}~\cite{fallas2023object}, frozen proposal classifiers~\cite{fallas2025squeeze}, and direct training-free detection~\cite{xie2024simple}. In contrast, our method operates directly on dense frozen features without an external proposal model or learned detection head.

\textbf{Precision Agriculture.}
\gls{od} supports fruit detection, crop monitoring, and yield estimation~\cite{sa2016deepfruits,kamilaris2018review}. Agricultural appearance varies across crops, locations, acquisition conditions, and growth stages, and object-level annotation is costly to repeat. Few-shot and training-free approaches are therefore attractive for adapting detectors to new agricultural settings.
\section{The Pineapple Plantation Dataset}
\label{sec:pineapple-dataset}

\begin{figure*}[t]
    \centering
    \includegraphics[width=\textwidth]{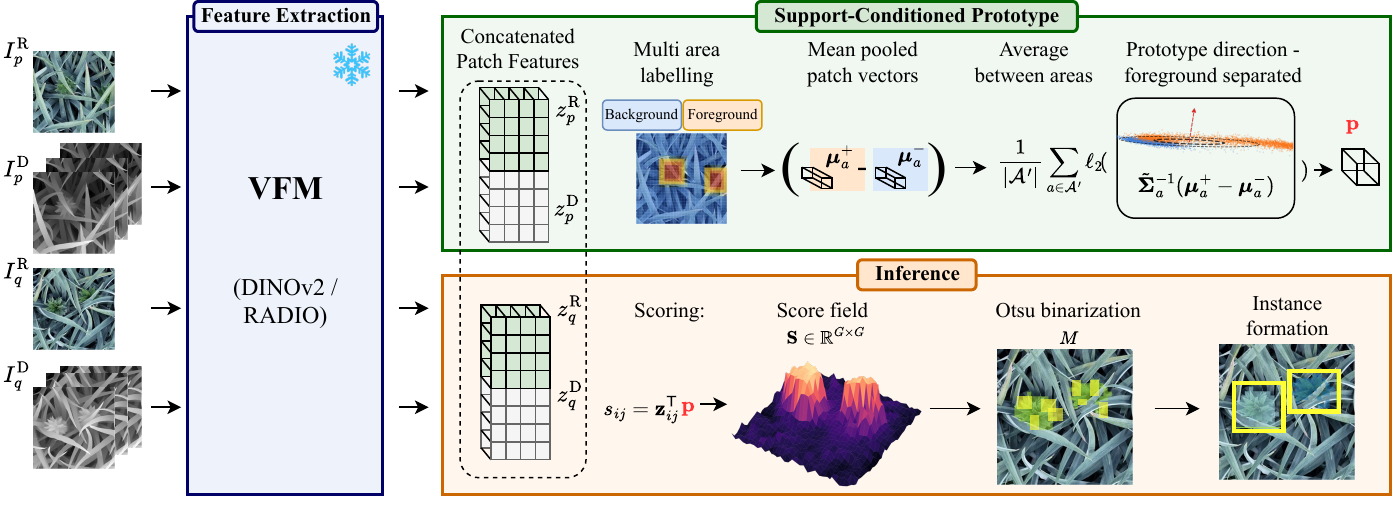}
    \caption{Overview of the proposed training-free detector. Support annotations define a foreground--background prototype from a frozen encoder. The prototype is applied to the query image to produce a score field through dot-product similarity; an adaptive Otsu threshold separates foreground, and support-derived geometric information converts the resulting regions into bounding boxes. No fine-tuning or learned detection head is required.}
    \label{fig:pipeline}
\end{figure*}

The Pineapple Plantation Dataset is a temporally structured benchmark for few-shot object detection and counting in dense agricultural scenes. In commercial plantations, pineapple counts support plant-loss estimation and harvest planning and are commonly obtained through manual field sampling. To emulate this process, low-altitude \gls{uav} flights were conducted at approximately 2\,m over a commercial pineapple plantation~\cite{11147940}, preserving the close-range view needed to distinguish individual crowns. The dataset contains 4,908 aerial \gls{rgb} frames, including 1,995 annotated frames with 17,609 pineapple bounding boxes, together with predicted relative-depth maps from Depth Anything v2~\cite{yang2024depthv2}. The imagery, annotations, depth maps, leakage-aware splits, and evaluation protocol are publicly available.

Annotations were manually created using Label Studio~\cite{Label_Studio} by trained annotators without model assistance. Plants were labeled if crowns were fully or partially visible with identifiable foliage, while severely occluded plants were excluded. Lead authors performed a final quality check to ensure accurate box alignment and eliminate errors.

To reduce temporal leakage, each flight is divided into 60-frame blocks assigned to training, validation, or test, discarding 20-frame buffers and cross-session revisits.  The resulting splits contain 3,536 training, 381 validation, and 346 test frames, including 1,500, 115, and 155 annotated frames with 14,097, 769, and 1,083 boxes, respectively. Evaluation uses fully annotated validation and test frames. Support images are sampled from distinct capture blocks to reduce the likelihood of observing the same plants across shots.

The dataset poses two challenges for counting: pineapple crowns have highly similar appearance to surrounding foliage, and they are frequently partially occluded by leaves, producing strong foreground--background ambiguity.

\section{Methodology}

\subsection{Problem Setting and Overview}
\label{sec:method-overview}

As illustrated in Fig.~\ref{fig:pipeline}, the proposed method follows a support-conditioned, training-free detection framework. An \emph{episode} consists of a support set
$\mathcal{S}_{K}=\{(I_n^{\mathrm{R}},I_n^{\mathrm{D}},\mathcal{B}_n)\}_{n=1}^{K}$,
where $I_n^{\mathrm{R}}$ denotes the \gls{rgb} image, $I_n^{\mathrm{D}}$ its aligned depth map, and $\mathcal{B}_n$ the corresponding object bounding-box annotations. The support set accompanies unannotated query images from the same domain, for which the goal is to predict a set of bounding boxes.

A frozen encoder maps each image to a $G\times G$ grid of $D$-dimensional patch tokens $\mathbf{z}_{ij}\in\mathbb{R}^{D}$; class and register tokens are discarded. The frozen encoders are \gls{dino}-B with registers at $448\times448$ and \gls{radio}~v2.5-B at $512\times512$, both producing a $32\times32$ token grid ($G=32$). Depth maps hold relative depth scaled per image to 8 bits, brighter is nearer (min--max for Depth Anything v2, 1st--99th percentile for Depth Anything 3); they are replicated across three channels and encoded exactly like \gls{rgb} images, including the encoder's input normalization. Tokens from each modality are $\ell_2$-normalized; for \gls{rgb}+D, the two token grids are concatenated and normalized again. The support annotations then define a prototype direction (Sec.~\ref{sec:prototype-construction}) that turns each query into a score field, which is thresholded per image (Sec.~\ref{sec:otsu-separation}) and converted into boxes with support-derived geometry (Sec.~\ref{sec:instance-decoding}); Fig.~\ref{fig:pipeline} summarizes these stages.

No gradient is computed, and no encoder or detector parameters are learned. All episode-specific quantities are estimated from the support set and refitted whenever it changes. The foreground threshold, however, is recomputed for every query image to account for image-to-image variation in illumination, plant density, and visible background.

\subsection{Support-Conditioned Prototype}
\label{sec:prototype-construction}

Patches near the center of an annotated box are more likely to describe the pineapple itself, whereas boundary patches often mix the plant with surrounding vegetation. To account for this variation, each ground-truth box defines concentric inner rectangles whose areas correspond to fractions $a\in\mathcal{A}=\{0.25,0.50,1.00\}$ of the original box. A patch whose center lies inside one of these rectangles is labelled as \emph{foreground} at scale $a$, forming $\mathcal{P}_{a}$. Patches outside the foreground rectangle but inside the corresponding box enlarged by a factor of $1.15$ are discarded as ambiguous. Patches outside all enlarged boxes form the \emph{background} set $\mathcal{N}$, shared across scales. The labelled descriptors at scale $a$ are
$\mathcal{L}_{a}=\mathcal{P}_{a}\cup\mathcal{N}$.

A simple prototype could use the mean difference
$\boldsymbol{\mu}^{+}_{a}-\boldsymbol{\mu}^{-}_{a}$.
However, pineapples and surrounding vegetation share strong visual characteristics, particularly leaf appearance, so directions with large variation in both groups can receive excessive importance. We therefore account for the covariance of the labelled descriptors. The direction is computed independently at each foreground scale, normalized, and averaged across scales:

\begin{equation}
    \mathbf{p}
      =\frac{1}{|\mathcal{A}'|}\sum_{a\in\mathcal{A}'}
        \operatorname{n}\!\left(
        \tilde{\boldsymbol{\Sigma}}_{a}^{-1}
        \left(\boldsymbol{\mu}^{+}_{a}-\boldsymbol{\mu}^{-}_{a}\right)\right),
    \label{eq:pyramid-prototype}
\end{equation}

where $\boldsymbol{\mu}^{\pm}_{a}$ are the means of $\mathcal{P}_{a}$ and $\mathcal{N}$, $\operatorname{n}(\cdot)$ denotes $\ell_2$ normalization, $\tilde{\boldsymbol{\Sigma}}_{a}$ is the regularized covariance of $\mathcal{L}_{a}$, and $\mathcal{A}'\subseteq\mathcal{A}$ retains the scales for which $\mathcal{P}_{a}$ is non-empty.

Because covariance is estimated from a limited number of support descriptors, poorly estimated directions can become overemphasized after inversion. We therefore regularize $\boldsymbol{\Sigma}_{a}$ by shrinkage toward a scaled identity matrix:

\begin{equation}
    \tilde{\boldsymbol{\Sigma}}_{a}
      =(1-\alpha_{a})\,\boldsymbol{\Sigma}_{a}
       +\alpha_{a}\,\frac{\operatorname{tr}\boldsymbol{\Sigma}_{a}}{D}\,\mathbf{I},
    \qquad
    \alpha_{a}=\frac{D}{D+|\mathcal{L}_{a}|}.
    \label{eq:covariance-shrinkage}
\end{equation}

The weight $\alpha_a$ depends only on support-set information: with fewer labelled descriptors, regularization is stronger; as more support descriptors become available, more weight is given to the estimated covariance.

Each query patch receives the scalar score
$s_{ij}=\mathbf{z}_{ij}^{\mathsf T}\mathbf{p}$,
measuring its alignment with the support-conditioned prototype. These scores form a field
$\mathbf{S}\in\mathbb{R}^{G\times G}$.

\subsection{Per-Image Foreground Separation}
\label{sec:otsu-separation}

The score field $\mathbf{S}$ indicates how strongly each query patch aligns with the prototype. Because illumination, vegetation density, and visible background vary across images, an image-specific threshold is estimated independently for every query image using Otsu's criterion~\cite{otsu1979threshold}.

Otsu partitions the scores into either $C=2$ or $C=3$ ordered groups. With $C=2$, the scores are separated into background and foreground. With $C=3$, an additional low-scoring background population, such as bare soil or sky, can be separated while preserving the highest-scoring group as foreground.

For either value of $C$, Otsu determines $C-1$ ordered boundaries
$t_1<\dots<t_{C-1}$. The highest boundary serves as the foreground threshold,

\begin{equation}
\tau_{\mathrm{O}} = t_{C-1},
\qquad
M_{ij}=\mathbf{1}\!\left[s_{ij}\geq\tau_{\mathrm{O}}\right],
\label{eq:otsu-mask}
\end{equation}

where $\mathbf{M}$ is the resulting binary foreground mask.

The choice between $C=2$ and $C=3$ is calibrated once on the support set at the beginning of each episode. Both alternatives are evaluated on the support images, and the configuration whose detections best reproduce the support annotations is retained, preferring $C=2$ in case of a tie. The selected $C$ remains fixed for the episode, while $\tau_{\mathrm{O}}$ is recomputed for every query image.

\subsection{From Regions to Pineapple Instances}
\label{sec:instance-decoding}

The binary mask identifies foreground patches but not individual pineapples: a single plant can form several disconnected regions, and some regions may still be false detections. The support set therefore provides object-scale and shape information for instance formation.

The median side length of the support boxes defines the typical object scale $\delta$ on the $G\times G$ patch grid. A shape cost compares each candidate region with the support-box geometries. Let $\mathbf{w}$ denote the width and height of a candidate region and $\mathbf{w}_{b}$ those of support box $b$:

\begin{equation}
d_{\mathcal{S}}(\mathbf{w})
=\min_{b\in\mathcal{B}_{\mathcal{S}}}
\left\lVert \log(\mathbf{w}/\mathbf{w}_{b}) \right\rVert_2 ,
\label{eq:shape-cost}
\end{equation}

where lower values indicate greater similarity to the support geometry.

Each connected foreground region is a candidate instance. Disconnected regions separated by fewer than $\delta$ cells are merged when their union provides a better geometric match than either region. The support-derived geometry therefore reconnects fragmented responses from the same plant.

Each resulting region $r$ is converted into a bounding box. Let $\mathbf{w}^{A}_r$ and $\mathbf{c}^{A}_r$ denote the size and centre of the box enclosing its foreground activation. Matched support regions provide a median size correction $\bar{\boldsymbol{\gamma}}$ and normalized centre offset $\bar{\boldsymbol{\Delta}}$:

\begin{align}
\widehat{\mathbf{w}}_r
&= \mathbf{w}^{A}_r \odot \bar{\boldsymbol{\gamma}},
\nonumber\\
\widehat{\mathbf{c}}_r
&= \mathbf{c}^{A}_r
+ \bar{\boldsymbol{\Delta}} \odot \widehat{\mathbf{w}}_r.
\label{eq:box-correction}
\end{align}

The corrected boxes are clipped to the image boundaries. Finally, candidate boxes are filtered using support-calibrated geometry and score prominence. The geometric acceptance threshold is selected to maximize support-set box F1.
A region $r$ that fails it is still kept when its score peak is prominent, i.e.\ when
\(
  \bigl(\max_{(i,j)\in r} s_{ij}-\tau_O\bigr)\big/\bigl(\max_{i,j} s_{ij}-\tau_O\bigr)
\)
reaches the 15th percentile of the same ratio over the support objects.
Thus, the final detections are obtained entirely from support-derived appearance and geometry, without query annotations.
\section{Experiments}
\label{sec:experiments}
\subsection{Protocol}
\label{sec:protocol}

\begin{table}[t]
  \centering
  \scriptsize
  \setlength{\tabcolsep}{4pt}
  \caption{\textbf{Annotation budget per support size.} Annotated boxes contained in the
  $K$ support images, i.e.\ the boxes every method receives, as the mean over the five
  paired support draws. Counts refer to the $544{\times}544$ support crops used by all
  methods. PhenoBench is evaluated up to $K{=}20$ since support images come from distinct capture blocks, and its training split has only $21$.}
  \label{tab:budget}
  \begin{tabular}{l*{6}{c}}
    \toprule
    Dataset & \textbf{1} & \textbf{2} & \textbf{4} & \textbf{10} & \textbf{20} & \textbf{40} \\
    \midrule
    Pineapple Plantation & 3 & 9 & 20 & 48 & 99 & 200 \\
    AmodalAppleSize      & 3 & 7 & 15 & 32 & 65 & 135 \\
    PhenoBench           & 9 & 21 & 39 & 87 & 172 & -- \\
    \bottomrule
  \end{tabular}
\end{table}

Every result is an average over five paired support draws. A draw fixes one episode, a set of $K$ support images and their boxes, and \emph{all} methods are scored on exactly that episode, so differences between methods never reflect differences between support sets. In each episode, $K$ support images are sampled from the training partition, and all held-out test images (PhenoBench: official validation set) serve as queries. Each support and query image is one fixed $544 \times 544$ crop: the central full-height crop of a $960 \times 544$ pineapple frame (keeping boxes centred inside it and at least $35$\% visible), the $1024 \times 1024$ content window of AmodalAppleSize, or the full PhenoBench image, both resized to $544$. No validation or test annotation contributes to prototype estimation, threshold configuration, geometric priors, or acceptance criteria. Support images come from distinct capture blocks, so no two shots observe the same plants; the Pineapple Plantation validation split does not contribute to reported results. The protocol evaluates $K\in\{1,2,4,10,20,40\}$. Table~\ref{tab:budget} lists the annotated boxes this gives ($3$ to $200$ on pineapple). 

Evaluation reports \gls{iou} F1 under maximum-cardinality one-to-one matching at \gls{iou}\,$\geq0.5$ (metric focused on localization), and count \gls{mae} (metric focused on counting correctly). Two external few-shot detectors serve as baselines, \gls{petdino}~\cite{Fu_2026_PETDINO} and \gls{geco2}~\cite{Pelhan_2026_GeCo2}, both run from released weights with no target-domain fine-tuning. \gls{geco2} takes its exemplars from the support images; following its release, each image is rescaled so that the mean exemplar side is $80$ pixels, and detections are kept above $0.33$ of the highest objectness score in the image. \gls{petdino} uses its default score threshold of $0.3$. Choosing their thresholds on the support images instead (leave one image out, no query labels) is unstable for $K \leq 2$ and helps from $K=4$, mostly for \gls{petdino} (pineapple \gls{rgb} $0.21\rightarrow0.48$ at $K=40$; PhenoBench \gls{rgb} $0.32\rightarrow0.51$ at $K=4$); \gls{geco2} changes by at most $0.08$. The pineapple ranking is unchanged at every $K$ and input, so we report the release defaults.  Because both are native \gls{rgb} models, their depth rows receive the predicted depth map replicated to three channels, and neither has an \gls{rgb}\,+\,D path. On one NVIDIA L40S (batch size 1), encoder passes take $7$--$8$\,ms (\gls{radio}/\gls{dino}-B) and Depth Anything v2 takes $40$\,ms; fitting a support set takes $0.2$--$9$\,s on four CPU threads ($K=1$--$40$), and decoding a query $2$--$16$\,ms. An \gls{rgb}+D query costs about $70$\,ms, against $0.1$--$0.8$\,s per query for \gls{petdino} and \gls{geco2}.

\begin{table*}[t]
  \centering
  \footnotesize
  \setlength{\tabcolsep}{4pt}
  \caption{\textbf{Pineapple test split, IoU F1.} $K$ denotes the number of annotated support images. Mean\,$\pm$\,standard deviation over five paired support draws; all methods receive identical episodes. Depth is predicted by Depth Anything~v2; the two competitors are native RGB models, so their depth rows feed the predicted depth map as a three-channel image. \textbf{Bold} marks the best cell per input condition and support size.}
  \label{tab:pineapple-iou}
  \begin{tabular}{llcccccc}
    \toprule
    Method & Input & \textbf{1} & \textbf{2} & \textbf{4} & \textbf{10} & \textbf{20} & \textbf{40} \\
    \midrule
    PET-DINO~\cite{Fu_2026_PETDINO} & RGB & 0.177\,$\pm$0.100 & 0.191\,$\pm$0.135 & 0.199\,$\pm$0.115 & 0.212\,$\pm$0.065 & 0.224\,$\pm$0.061 & 0.210\,$\pm$0.023 \\
    PET-DINO~\cite{Fu_2026_PETDINO} & Depth & 0.142\,$\pm$0.100 & 0.153\,$\pm$0.117 & 0.095\,$\pm$0.076 & 0.091\,$\pm$0.051 & 0.079\,$\pm$0.026 & 0.071\,$\pm$0.019 \\
    \midrule
    GeCo2~\cite{Pelhan_2026_GeCo2} & RGB & 0.319\,$\pm$0.166 & 0.328\,$\pm$0.070 & 0.339\,$\pm$0.049 & 0.362\,$\pm$0.045 & 0.373\,$\pm$0.026 & 0.371\,$\pm$0.018 \\
    GeCo2~\cite{Pelhan_2026_GeCo2} & Depth & 0.204\,$\pm$0.157 & 0.188\,$\pm$0.085 & 0.178\,$\pm$0.056 & 0.158\,$\pm$0.025 & 0.147\,$\pm$0.014 & 0.138\,$\pm$0.003 \\
    \midrule
    \rowcolor{gray!15}
    Ours (AM-RADIO) & RGB & 0.318\,$\pm$0.104 & 0.442\,$\pm$0.061 & 0.481\,$\pm$0.057 & 0.520\,$\pm$0.020 & 0.485\,$\pm$0.036 & 0.530\,$\pm$0.027 \\
    \rowcolor{gray!15}
    Ours (AM-RADIO) & Depth & 0.250\,$\pm$0.130 & 0.372\,$\pm$0.052 & 0.439\,$\pm$0.064 & 0.503\,$\pm$0.029 & 0.462\,$\pm$0.023 & 0.495\,$\pm$0.022 \\
    \rowcolor{gray!15}
    Ours (AM-RADIO) & RGB+D & 0.322\,$\pm$0.046 & 0.466\,$\pm$0.048 & 0.499\,$\pm$0.053 & 0.555\,$\pm$0.017 & 0.511\,$\pm$0.039 & 0.563\,$\pm$0.024 \\
    \midrule
    \rowcolor{gray!15}
    Ours (DINOv2-B) & RGB & \textbf{0.349}\,$\pm$0.130 & \textbf{0.513}\,$\pm$0.072 & \textbf{0.530}\,$\pm$0.043 & \textbf{0.558}\,$\pm$0.043 & \textbf{0.554}\,$\pm$0.041 & \textbf{0.573}\,$\pm$0.028 \\
    \rowcolor{gray!15}
    Ours (DINOv2-B) & Depth & \textbf{0.317}\,$\pm$0.096 & \textbf{0.463}\,$\pm$0.060 & \textbf{0.500}\,$\pm$0.037 & \textbf{0.542}\,$\pm$0.017 & \textbf{0.511}\,$\pm$0.033 & \textbf{0.529}\,$\pm$0.017 \\
    \rowcolor{gray!15}
    Ours (DINOv2-B) & RGB+D & \textbf{0.362}\,$\pm$0.108 & \textbf{0.541}\,$\pm$0.037 & \textbf{0.555}\,$\pm$0.013 & \textbf{0.567}\,$\pm$0.011 & \textbf{0.546}\,$\pm$0.033 & \textbf{0.565}\,$\pm$0.016 \\
    \bottomrule
  \end{tabular}
\end{table*}

\begin{figure}[t]
  \centering
  \includegraphics[width=0.95\columnwidth]{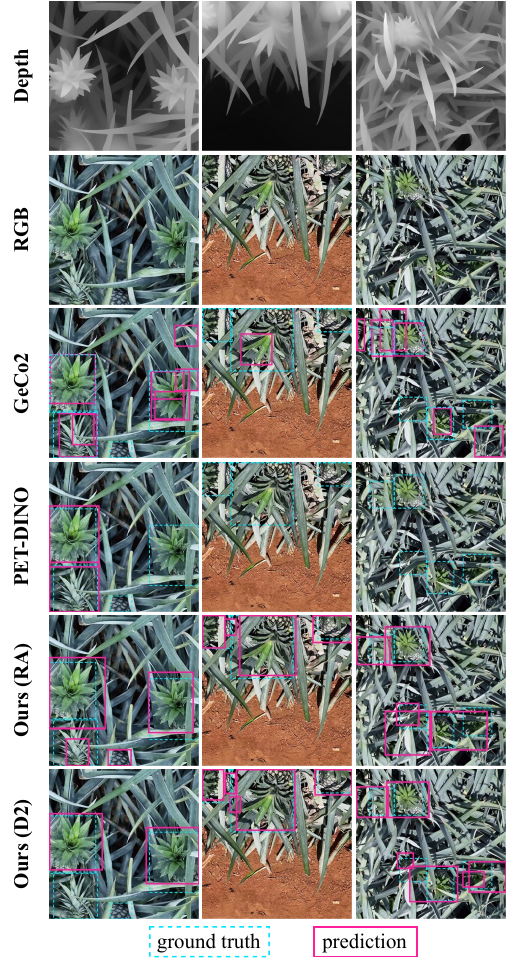}
  \caption{Qualitative comparison on the Pineapple Plantation Dataset. Predictions from \gls{geco2}, \gls{petdino}, and the proposed method with \gls{radio} (RA) and \gls{dino}-B (D2) are shown against the ground truth. All the methods used $K$=40.}
  \label{fig:pineapple_qualitative}
\end{figure}
\subsection{Comparison with few-shot detectors}
\label{sec:comparison}

Table~\ref{tab:pineapple-iou} reports the evaluation results. With \gls{dino}-B, the proposed detector leads at every support size and in every input condition, by $1.1\times$ at $K=1$ and by $1.5\times$ at $K=40$ against the stronger competitor; \gls{radio} matches \gls{geco2} on \gls{rgb} at $K=1$ ($0.318$ vs. $0.319$) and leads elsewhere. The curves differ in shape as well as in levels: the detector performance rises by $64\%$ between one and forty support images, \gls{petdino} plateaus near $0.21$, and \gls{geco2} rises only $16\%$, from $0.319$ to $0.371$. Figure~\ref{fig:pineapple_qualitative} shows predictions at $K=40$.

\begin{figure}[t]
  \centering
  \includegraphics[width=0.9\columnwidth]{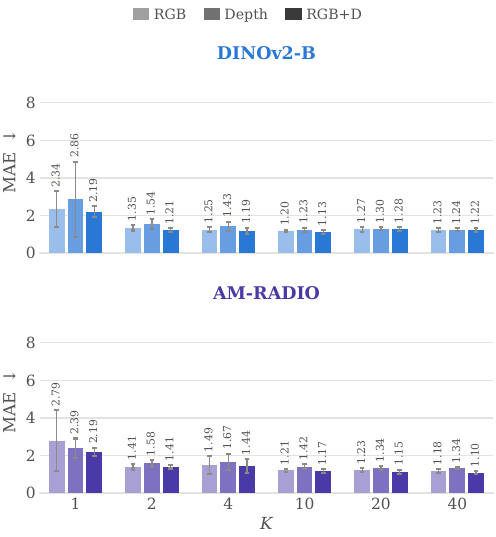}
  \caption{Counting error of the proposed detector on the pineapple test split against support size $K$, for both frozen backbones and all three input conditions. Bars are the mean over 5 seeds and whiskers are $\pm$1 s.d. Both baselines are worse than every bar and are omitted from the axis so that the conditions remain legible: over $K$=1 to $40$, \gls{petdino} ranges $3.03$--$3.16$ (\gls{rgb}) and $2.88$--$3.34$ (depth), and \gls{geco2} ranges $2.96$--$3.25$ (\gls{rgb}) and $9.95$--$16.3$ (depth); for $K\geq2$ their \gls{rgb} error is two to three times ours.}
  \label{fig:pineapple_mae_shots}
\end{figure}

A slightly different trend can be observed for counting-based metrics such as \gls{mae}, as shown in Figure~\ref{fig:pineapple_mae_shots}. At $K=40$, \gls{radio} achieves a lower \gls{mae} than \gls{dino}-B with the \gls{rgb}+D modality, whereas \gls{dino}-B is lower at $K=2$--$10$ and the two are tied at $K=1$. Another important observation is that, for both models, \gls{rgb}+D improves over \gls{rgb}-only in terms of \gls{mae}, whereas the same benefit is consistent only for \gls{radio} for the \gls{iou} F1 score. In contrast, neither baseline counts better with more support: their \gls{rgb} \gls{mae} stays near $3$, and their depth \gls{mae} grows with $K$. The deterioration is particularly relevant for the Pineapple Plantation Dataset, where accurate pineapple counting is an important agricultural application.

\begin{table}[t]
  \centering
  \scriptsize
  \setlength{\tabcolsep}{2pt}
  \caption{\textbf{Shuffled-depth control}, RGB\,+\,D, pineapple test. Each query keeps its RGB features but takes depth from another capture block, destroying only spatial correspondence. $\Delta$ is signed so positive favours aligned depth (aligned $-$ shuffled for F1, shuffled $-$ aligned for MAE). \emph{mean} and \emph{wins} span all six support sizes, including $K{=}2$ and $10$, not shown.}
  \label{tab:shuffle}
  \begin{tabular}{llcccccc}
    \toprule
    Encoder & Depth & \textbf{1} & \textbf{4} & \textbf{20} & \textbf{40} & \textbf{mean} & \textbf{wins} \\
    \midrule
    \multicolumn{8}{l}{\emph{IoU F1 $\uparrow$}} \\[1pt]
    DINOv2-B & aligned & 0.362 & 0.555 & 0.546 & 0.565 & & \\
     & shuffled & 0.370 & 0.535 & 0.524 & 0.579 & & \\
     & $\Delta$ & $-0.008$ & $+0.020$ & $+0.022$ & $-0.013$ & $\mathbf{+0.009}$ & 3/6 \\
    \addlinespace[1pt]
    AM-RADIO & aligned & 0.322 & 0.499 & 0.511 & 0.563 & & \\
     & shuffled & 0.302 & 0.437 & 0.488 & 0.538 & & \\
     & $\Delta$ & $+0.020$ & $+0.062$ & $+0.023$ & $+0.025$ & $\mathbf{+0.038}$ & 6/6 \\
    \midrule
    \multicolumn{8}{l}{\emph{count MAE $\downarrow$}} \\[1pt]
    DINOv2-B & aligned & 2.19 & 1.19 & 1.28 & 1.22 & & \\
     & shuffled & 2.80 & 1.28 & 1.33 & 1.21 & & \\
     & $\Delta$ & $+0.60$ & $+0.09$ & $+0.05$ & $-0.01$ & $\mathbf{+0.18}$ & 5/6 \\
    \addlinespace[1pt]
    AM-RADIO & aligned & 2.19 & 1.44 & 1.15 & 1.10 & & \\
     & shuffled & 2.48 & 1.81 & 1.24 & 1.15 & & \\
     & $\Delta$ & $+0.29$ & $+0.37$ & $+0.09$ & $+0.06$ & $\mathbf{+0.18}$ & 6/6 \\
    \bottomrule
  \end{tabular}
\end{table}

A shuffled-depth control tests whether fusion benefits from aligned geometry rather than merely from a second feature stream (Table~\ref{tab:shuffle}). Query-depth features are shuffled across capture blocks while preserving their distribution. Alignment improves counting \gls{mae} for both encoders and \gls{iou} F1 consistently for \gls{radio}, showing that spatially corresponding depth carries useful instance information.

\subsection{Component ablation and transfer domain}
\label{sec:ablation}

\begin{table}[htbp]
  \centering
  \scriptsize
      \caption{\textbf{Component ablation} (\gls{dino}-B, \gls{rgb}, pineapple test). IoU F1 over five paired draws. (a): neither contribution; (d): full method; (e): fixed $\alpha=0.15$ instead of token-to-dimension shrinkage. Bold: best per support size; shaded: final setting.}
  \label{tab:ablation}

  \begin{tabularx}{\columnwidth}{
        l
        l
        l
        *{4}{>{\centering\arraybackslash}X}
    }
      \toprule
      & Prototype & Otsu & \textbf{1} & \textbf{4} & \textbf{20} & \textbf{40} \\
      \midrule
      (a) & cosine & $C{=}2$ & 0.335 & 0.492 & 0.470 & 0.482 \\
      (b) & cosine & fitted & \textbf{0.369} & 0.517 & 0.506 & 0.520 \\
      (c) & whitened & $C{=}2$ & 0.349 & 0.505 & 0.516 & 0.538 \\
      \rowcolor{gray!15}
      (d) & whitened & fitted & 0.349 & \textbf{0.530} & \textbf{0.554} & \textbf{0.573} \\
      (e) & whitened$^{\alpha}$ & fitted & 0.281 & 0.525 & 0.553 & 0.572 \\
      \bottomrule
    \end{tabularx}
\end{table}

Two elements distinguish the proposed prototype and threshold from a plain mean-difference prototype with a two-class Otsu cut: whitening by the support covariance, and fitting the number of score populations on the support set. Table~\ref{tab:ablation} adds them one at a time to the same pipeline. They contribute, and they compose. Fitting the class count alone adds $+0.038$ \gls{iou} F1 at $K=40$, whitening alone $+0.056$, and together $+0.091$. The gain from whitening increases with support, from
$+0.014$ at $K=1$ to $+0.056$ at $K=40$, because the covariance it inverts is better estimated as tokens accumulate, whereas the fitted class count helps right
from the first shot. Row~(e) isolates the shrinkage rule: replacing the token-to-dimension weight with a fixed $\alpha{=}0.15$ costs $0.068$ \gls{iou} F1 at $K=1$. This is the regime where a one-image covariance is underdetermined and a fixed weight under-regularises it.

\begin{figure}[t]
  \centering
  \includegraphics[width=\columnwidth]{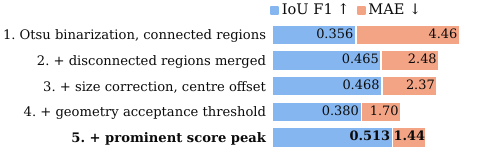}
  \caption{\textbf{Decoder ablation}, \gls{dino}-B, \gls{rgb}, pineapple test. Each row adds one stage of Section~\ref{sec:instance-decoding}; the bold row is the full decoder. Bars show \gls{iou} F1 (blue,$\uparrow$) and count \gls{mae} (orange, $\downarrow$), averaged over five draws and all $K$; each segment has its own scale.}
  \label{fig:ablation_Decoding}
\end{figure}

The instance decoder is ablated in the same way (Figure~\ref{fig:ablation_Decoding}): starting from the connected regions of the Otsu mask, each support-derived stage of Section~\ref{sec:instance-decoding} is added in turn (\gls{iou} F1 and \gls{mae} averaged over $K$). Merging disconnected regions gives the largest gain ($0.356$ to $0.465$ \gls{iou} F1, $4.46$ to $2.48$ \gls{mae}), and the size and centre correction adds little. The geometry acceptance threshold alone lowers \gls{iou} F1 to $0.380$ while reducing \gls{mae}, because it also rejects true detections; the prominent-peak criterion recovers them, reaching $0.513$ \gls{iou} F1 and $1.44$ \gls{mae}. Table~\ref{tab:encoder} isolates the frozen encoder: replacing \gls{dino}-B with an ImageNet-supervised ResNet-50 lowers \gls{iou} F1 at every support size, yet on pineapple the same pipeline still outperforms the strongest baseline from $K{=}4$ onward (0.493 vs.\ 0.371 at $K{=}40$), so the gains do not come from the foundation-model features alone.

\begin{table}[t]
  \centering
  \scriptsize
  \setlength{\tabcolsep}{4pt}
  \caption{\textbf{Frozen-encoder ablation (\gls{rgb} IoU F1).} Mean over five paired support draws. Our pipeline uses \gls{dino}-B or ImageNet-supervised ResNet-50~\cite{He_ResNet_2016} (layer 3, $32\times32$ grid) without modification. The strongest baseline per dataset is included; bold marks the best score per dataset and support size.}
  \label{tab:encoder}
  \begin{tabular}{@{}l*{4}{c}@{}}
    \toprule
    Method & \textbf{1} & \textbf{4} & \textbf{20} & \textbf{40} \\
    \midrule
    \multicolumn{5}{@{}l}{\textit{Pineapple (aerial)}} \\
    \quad Ours, DINOv2-B & \textbf{0.349} & \textbf{0.530} & \textbf{0.554} & \textbf{0.573} \\
    \quad Ours, ResNet-50 & 0.164 & 0.351 & 0.436 & 0.493 \\
    \quad GeCo2 (strongest baseline) & 0.319 & 0.339 & 0.373 & 0.371 \\
    \bottomrule
  \end{tabular}
\end{table}

\newcommand{\val}[2]{#1{\tiny$\pm$#2}}
\newcommand{\best}[2]{\textbf{#1}{\tiny$\pm$#2}}

\begin{table}[t]
  \centering
  \scriptsize
  \setlength{\tabcolsep}{2pt}
  \renewcommand{\arraystretch}{1.05}

  \caption{\textbf{The same protocol on three benchmarks.}
  IoU F1, mean over five paired support draws, with standard deviation
  shown in smaller type. \emph{Ours (D2)} and \emph{Ours (RA)} correspond to
  \gls{dino}-B and \gls{radio}, respectively. Depth: Depth Anything v2~\cite{yang2024depthv2} (pineapple), Depth Anything 3~\cite{lin2026depthanythingV3} (others). Dashes denote
  unavailable \gls{rgb}\,+\,D settings. \textbf{Bold} indicates the best result within each dataset and column. Shaded rows denote the proposed methods.}
  \label{tab:other-datasets}

  \begin{tabular}{@{}l*{6}{c}@{}}
    \toprule
    & \multicolumn{2}{c}{RGB}
    & \multicolumn{2}{c}{Depth}
    & \multicolumn{2}{c}{RGB\,+\,D} \\
    \cmidrule(lr){2-3}
    \cmidrule(lr){4-5}
    \cmidrule(lr){6-7}

    Method & $K$4 & $K$20 & $K$4 & $K$20 & $K$4 & $K$20 \\
    \midrule

    \multicolumn{7}{@{}l}{
      \shortstack[l]{
        \textit{Pineapple (aerial)}
      }
    } \\[-1pt]

    PET-DINO
      & \val{0.199}{.11}
      & \val{0.224}{.06}
      & \val{0.095}{.07}
      & \val{0.079}{.02}
      & -- & -- \\

    GeCo2
      & \val{0.339}{.04}
      & \val{0.373}{.02}
      & \val{0.178}{.05}
      & \val{0.147}{.01}
      & -- & -- \\[1pt]
    \rowcolor{gray!15}
    Ours (D2)
      & \best{0.530}{.04}
      & \best{0.554}{.04}
      & \best{0.500}{.03}
      & \best{0.511}{.03}
      & \best{0.555}{.01}
      & \best{0.546}{.03} \\
    \rowcolor{gray!15}
    Ours (RA)
      & \val{0.481}{.05}
      & \val{0.485}{.03}
      & \val{0.439}{.06}
      & \val{0.462}{.02}
      & \val{0.499}{.05}
      & \val{0.511}{.03} \\

    \midrule

    \multicolumn{7}{@{}l}{
      \shortstack[l]{
        \textit{AmodalAppleSize (ground-level)}
      }
    } \\[-1pt]

    PET-DINO
      & \best{0.749}{.01}
      & \best{0.769}{.00}
      & \best{0.285}{.03}
      & \val{0.214}{.04}
      & -- & -- \\

    GeCo2
      & \val{0.644}{.04}
      & \val{0.642}{.01}
      & \val{0.173}{.02}
      & \val{0.170}{.00}
      & -- & -- \\[1pt]
    \rowcolor{gray!15}
    Ours (D2)
      & \val{0.414}{.06}
      & \val{0.442}{.04}
      & \val{0.237}{.04}
      & \best{0.245}{.04}
      & \val{0.410}{.05}
      & \best{0.446}{.03} \\
    \rowcolor{gray!15}
    Ours (RA)
      & \val{0.402}{.06}
      & \val{0.437}{.04}
      & \val{0.156}{.03}
      & \val{0.209}{.01}
      & \best{0.415}{.03}
      & \val{0.441}{.03} \\

    \midrule

    \multicolumn{7}{@{}l}{
      \shortstack[l]{
        \textit{PhenoBench (aerial)}
      }
    } \\[-1pt]

    PET-DINO
      & \val{0.317}{.11}
      & \val{0.228}{.04}
      & \val{0.515}{.07}
      & \best{0.476}{.03}
      & -- & -- \\

    GeCo2
      & \best{0.518}{.01}
      & \best{0.480}{.00}
      & \best{0.534}{.03}
      & \val{0.475}{.02}
      & -- & -- \\[1pt]
    \rowcolor{gray!15}
    Ours (D2)
      & \val{0.369}{.02}
      & \val{0.422}{.01}
      & \val{0.345}{.07}
      & \val{0.370}{.02}
      & \val{0.339}{.05}
      & \val{0.434}{.01} \\
    \rowcolor{gray!15}
    Ours (RA)
      & \val{0.318}{.02}
      & \val{0.406}{.01}
      & \val{0.358}{.06}
      & \val{0.389}{.01}
      & \best{0.378}{.03}
      & \best{0.437}{.01} \\

    \bottomrule
  \end{tabular}
\end{table}

The transfer beyond the aerial pineapple domain is evaluated under the same protocol on AmodalAppleSize~\cite{GENEMOLA2024AmodalAppleSize}, a ground-level \gls{rgb}+D orchard dataset with heavy occlusion, and on PhenoBench~\cite{Weyler2024PhenoBench}, an aerial crop/weed benchmark with crops as the target class. PhenoBench has no depth, and the sensor depth released with AmodalAppleSize is valid for only 64\% of pixels, so both use depth predicted by Depth Anything~3~\cite{lin2026depthanythingV3} (Table~\ref{tab:other-datasets}), whereas pineapple keeps the Depth Anything v2~\cite{yang2024depthv2} maps released with its dataset. Recomputing the pineapple depth with Depth Anything~3 changes \gls{rgb}+D \gls{iou} F1 by 0.00 (\gls{dino}-B) and $-0.02$ (\gls{radio}) on average.
The resulting transfer remains domain-dependent. On PhenoBench \gls{rgb}, \gls{geco2} is the strongest method at $K$=4 and $K$=20, but the proposed detector is the only evaluated method whose \gls{rgb} performance improves substantially with additional support ($0.369$ to $0.422$ for \gls{dino}-B, while \gls{geco2} drops from $0.518$ to $0.480$). More importantly for the aerial setting, depth alone preserves roughly 90\% or more of the \gls{rgb} \gls{iou} F1 of both encoders on Pineapple and PhenoBench, despite discarding appearance. \gls{rgb}+D also often improves over \gls{rgb} alone; on PhenoBench at $K$=20, fusion improves both encoders ($0.434$ and $0.437$), narrowing the gap to \gls{geco2}. Collectively, the results show that geometry offers a complementary signal for aerial imagery. 

Dataset-specific object statistics explain the ranking differences. Pineapple crowns are relatively uniform in size (median $6.9$ grid cells; $2.6\times$ variation), so support-derived geometry transfers well across queries. AmodalAppleSize is harder because apples are heavily occluded and annotated with amodal boxes, while PhenoBench has much larger scale variation ($13.9\times$) and many sub-cell objects ($24\%$), making a single support-derived geometry unreliable. \gls{geco2} benefits there from exemplar rescaling. Conversely, \gls{petdino} and \gls{geco2} are weakest on pineapple, where visually similar surrounding foliage makes prompting alone less effective. Our method benefits from re-estimating the prototype, threshold, and geometry from each support set.

\section{Conclusion}
\label{sec:conclusion}

We presented a training-free detector that converts dense features from frozen visual foundation models directly into detections, without fine-tuning, a learned head, or external proposals. On our leakage-aware Pineapple Plantation Dataset, it improves with additional support and reaches $1.5\times$ the \gls{iou} F1 of the strongest baseline at $K$=40. With \gls{dino}-B, predicted depth alone retains 92\% of \gls{rgb} \gls{iou} F1 at $K$=40, while \gls{rgb}+depth fusion improves counting \gls{mae} for both encoders. Experiments on PhenoBench and AmodalAppleSize show that transfer is domain-dependent and sensitive to depth quality: the proposed detector is strongest when objects span several feature-grid cells and share a consistent geometry, as pineapple crowns do, while the prompt-based baselines remain stronger for heavily occluded fruit and for plants of widely varying size. These results establish frozen dense representations as a promising basis for annotation-efficient agricultural detection.

{
    \small
    \bibliographystyle{ieeenat_fullname}
    \bibliography{main}
}

\clearpage
\appendix
\maketitlesupplementary

\section{AM-RADIO heads}
\label{sec:radio_spaces}
\begin{figure}[htbp]
  \centering
  \includegraphics[width=\columnwidth]{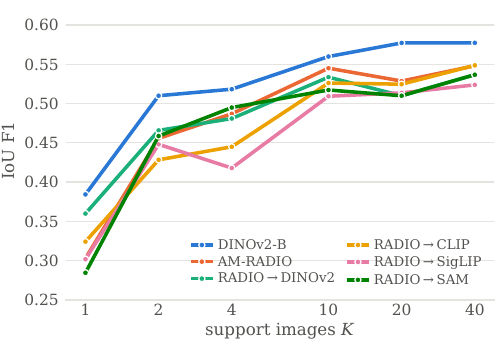}
  \caption{\textbf{Six frozen readout spaces}, RGB, pineapple test. The plain
  DINOv2-B backbone leads at every support size; AM-RADIO's distilled teacher heads neither help nor collapse, spanning 0.51--0.55 at $K$=40.}
  \label{fig:readout-spaces}
\end{figure}

Our pipeline is agnostic to the choice of frozen encoder; accordingly, we evaluate it across six readout spaces: DINOv2-B, the AM-RADIO backbone, and AM-RADIO's four distilled teacher heads (Figure~\ref{fig:readout-spaces}). DINOv2-B achieves the highest IoU F1 score at every support size, while the four teacher heads fall within a very small margin of the AM-RADIO backbone at $K=40$. These results are consistent with the nature of these representation spaces. The AM-RADIO backbone jointly encodes information distilled from the four teacher spaces, which may explain why the individual teacher heads perform slightly below it in this use case.

\begin{figure}[t]
  \centering
  \includegraphics[width=\linewidth]{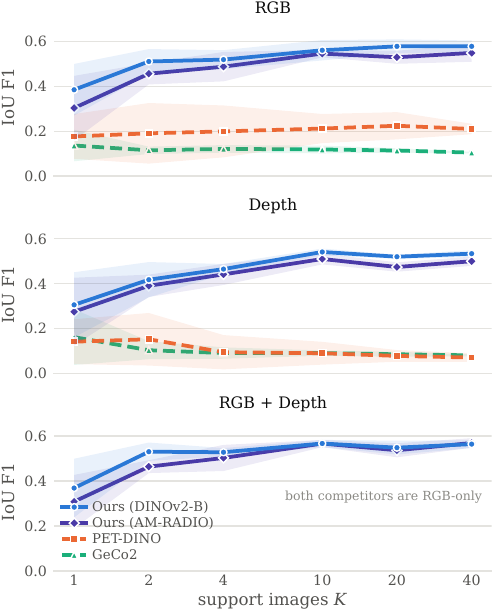}
  \caption{\textbf{Pineapple test split.} Detection quality (IoU F1, higher is better) against the number of support images for RGB, depth, and RGB+depth inputs. Bands are $\pm$1 standard deviation over the five paired draws; the proposed detector appears on both frozen encoders (solid) against the two competitors (dashed). Performance generally increases with additional support, although the gains are not monotonic for every encoder and input condition. DINOv2-B rises from $0.349$ to $0.573$ IoU F1 on RGB, whereas PET-DINO is flat and GeCo2 improves only slightly on RGB ($0.319$ to $0.371$) and degrades on depth.}
  \label{fig:pineapple-main}
\end{figure}

\textbf{Predicted depth alone is nearly as good as RGB for F1 score, and for MAE with DINOv2-B it stays within $0.04$ of RGB for $K\geq10$.} Depth reaches 0.529 against 0.573 at $K$=40, and 92--97\% of the RGB score for every $K\geq4$ in Table~\ref{tab:pineapple-iou} (see also Figure~\ref{fig:pineapple-main}). Although this is a monocular estimate produced from the same image rather than a measurement, a detector built on it alone recovers almost all of the accuracy. The estimator evidently concentrates the crown/canopy boundary into a channel where it survives our prototype and threshold better than the raw appearance does. For MAE, Figure~\ref{fig:pineapple_mae_shots} shows that depth alone stays close to RGB for DINOv2-B but trails it for AM-RADIO. RGB+D gives the lowest MAE of the three inputs at every $K$ except $K=20$ for DINOv2-B and except $K=2$ for AM-RADIO, and at $K=40$ AM-RADIO with RGB+D gives the lowest MAE of both encoders ($1.10$).

\begin{figure}[htbp]
  \centering
  \includegraphics[width=\columnwidth]{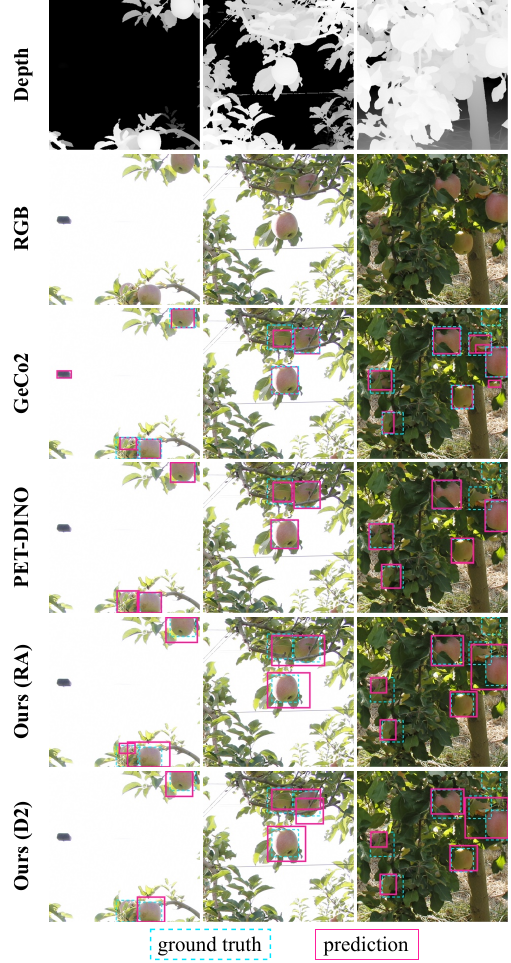}
  \caption{Qualitative comparison on the AmodalAppleSize~\cite{GENEMOLA2024AmodalAppleSize} dataset. Predictions from GeCo2, PET-DINO, and the proposed method with AM-RADIO (RA) and DINOv2-B (D2) are shown against the ground truth. All the methods used $K$=20.}
  \label{fig:apple_qualitative}
\end{figure}

\begin{figure}[htbp]
  \centering
  \includegraphics[width=\columnwidth]{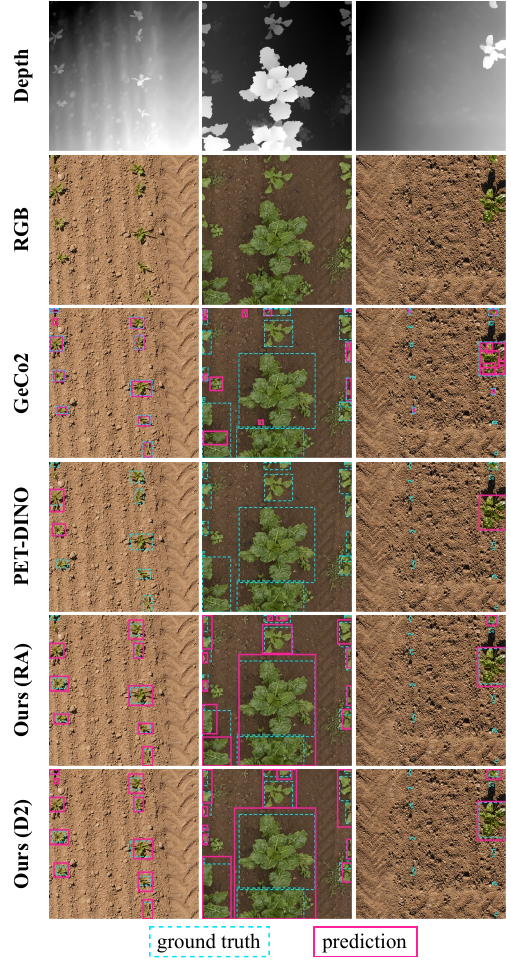}
  \caption{Qualitative comparison on the PhenoBench~\cite{Weyler2024PhenoBench} dataset. Predictions from GeCo2, PET-DINO, and the proposed method with AM-RADIO (RA) and DINOv2-B (D2) are shown against the ground truth. All the methods used $K$=20.}
  \label{fig:pheno_qualitative}
\end{figure}

\end{document}